\documentclass[11pt,a4paper]{article}
\usepackage[hyperref]{acl2021}
\usepackage{times}
\usepackage{latexsym}

\newcommand\figlength{4.5cm}

\usepackage{microtype}
\usepackage{enumerate}
\usepackage{graphicx}
\usepackage{makecell}
\usepackage{booktabs}
\usepackage{multirow}
\usepackage{xcolor}
\usepackage{subcaption}

\aclfinalcopy 

\title{An Exploratory Analysis of Multilingual Word-Level Quality Estimation with Cross-Lingual Transformers}

\author{Tharindu Ranasinghe$^\diamondsuit$, \textbf{Constantin Or\u{a}san$^\heartsuit$ and Ruslan Mitkov$^\diamondsuit$} \\
 $^\diamondsuit$Research Group in Computational Linguistics, University of Wolverhampton, UK \\
 $^\heartsuit$Centre for Translation Studies, University of Surrey, UK \\
 {\tt \{t.d.ranasinghehettiarachchige, r.mitkov\}@wlv.ac.uk} \\
 {\tt  c.orasan@surrey.ac.uk} }

\date{}

\begin{document}
\maketitle
\begin{abstract}
Most studies on word-level Quality Estimation (QE) of machine translation focus on language-specific models. The obvious disadvantages of these approaches are the need for labelled data for each language pair and the high cost required to maintain several language-specific models. To overcome these problems, we explore different approaches to multilingual, word-level QE. We show that these QE models perform on par with the current language-specific models. In the cases of zero-shot and few-shot QE, we demonstrate that it is possible to accurately predict word-level quality for any given new language pair from models trained on other language pairs. Our findings suggest that the word-level QE models based on powerful pre-trained transformers that we propose in this paper generalise well across languages, making them more useful in real-world scenarios.
\end{abstract}

\section{Introduction}
\label{sec:intro}
Quality Estimation (QE) is the task of assessing the quality of a translation without having access to a reference translation \cite{specia-etal-2009-estimating}. Translation quality can be estimated at different levels of granularity: word, sentence and document level \cite{ive-etal-2018-deepquest}. So far the most popular task has been sentence-level QE \cite{specia-etal-2020-findings-wmt}, in which QE models provide a score for each pair of source and target sentences. A more challenging task, which is currently receiving a lot of attention from the research community, is word-level quality estimation. This task provides more fine-grained information about the quality of a translation, indicating which words from the source have been incorrectly translated in the target, and whether the words inserted between these words are correct (good vs bad gaps).
This information can be useful for post-editors by indicating the parts of a sentence on which they have to focus more.

Word-level QE is generally framed as a supervised 
ML problem \cite{kepler-etal-2019-openkiwi,lee-2020-two} trained on data in which the correctness of translation is labelled at word level (i.e. good, bad, gap). 
The training data publicly available to build word-level QE models is limited to very few language pairs, which makes it difficult to build QE models for many languages.
From an application perspective, even for the languages with resources, it is difficult to maintain separate QE models for each language 
since the state-of-the-art neural QE models are large in size \cite{ranasinghe-etal-2020-transquest}. 

\begin{figure*}
\centering
\includegraphics[scale=0.57]{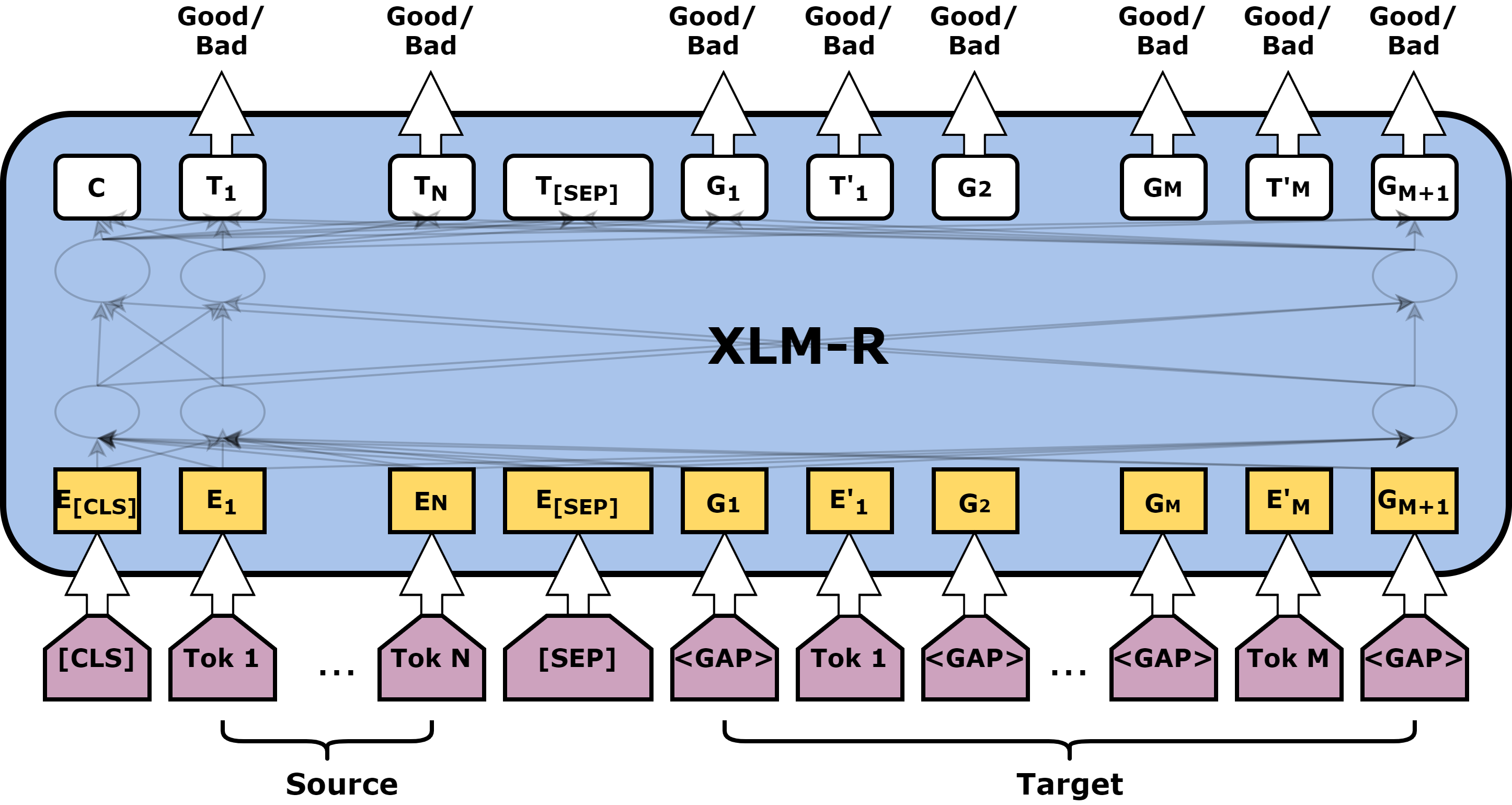}
\caption{Model Architecture}
\label{fig:architecture}
\end{figure*}

In our paper, we address this problem by developing multilingual word-level QE models which perform competitively in different domains, MT types and language pairs. In addition, for the first time, we propose word-level QE as a zero-shot cross-lingual transfer task, enabling new avenues of research in which multilingual models can be trained once and then serve a multitude of languages and domains. The main contributions of this paper are the following:

\begin{enumerate}[i]
  \vspace{-2mm}
  \item We introduce a simple architecture to perform word-level quality estimation that predicts the quality of the words in the source sentence, target sentence and the gaps in the target sentence.
  \vspace{-6mm}
  \item We explore multilingual, word-level quality estimation with the proposed architecture. We show that multilingual models are competitive with bilingual models.
  \vspace{-2mm}
  \item We inspect few-shot and zero-shot word-level quality estimation with the bilingual and multilingual models. We report how the source-target direction, domain and MT type affect the predictions for a new language pair.
  \vspace{-2mm}
  \item We release the code and the pre-trained models as part of an open-source framework\footnote{Documentation is available on \url{http://tharindu.co.uk/TransQuest/}}.

\end{enumerate}

\section{Related Work}
\label{sec:related}
\paragraph{QE}
Early approaches in word-level QE were based on features fed into a traditional machine learning algorithm. Systems like QuEst++ \cite{specia-etal-2015-multi} and MARMOT \cite{logacheva-etal-2016-marmot} were based on features used with Conditional Random Fields to perform word-level QE. With deep learning models becoming popular, the next generation of word-level QE algorithms were based on bilingual word embeddings fed into deep neural networks.  Such approaches can be found in OpenKiwi \cite{kepler-etal-2019-openkiwi}. However, the current state of the art in word-level QE is based on transformers like BERT \cite{devlin-etal-2019-bert} and XLM-R \cite{conneau-etal-2020-unsupervised} where a simple linear layer is added on top of the transformer model to obtain the predictions \cite{lee-2020-two}. All of these approaches consider quality estimation as a language-specific task and build a different model for each language pair. This approach has many drawbacks in real-world applications, some of which are discussed in Section \ref{sec:intro}.

\paragraph{Multilinguality}
Multilinguality allows training a single model to perform a task from and/or to multiple languages. Even though this has been applied to many tasks \cite{ranasinghe-zampieri-2020-multilingual,ranasinghe-zampieri-2021-mudes} including NMT \cite{nguyen-chiang-2017-transfer, aharoni-etal-2019-massively}, multilingual approaches have been rarely used in QE \cite{sun-etal-2020-exploratory}. \citet{shah-specia-2016-large} explore QE models for more than one language where they use multitask learning with annotators or languages as multiple tasks. They show that multilingual models led to marginal improvements over bilingual ones with a traditional black-box, feature-based approach. In a recent study, \citet{ranasinghe-etal-2020-transquest} show that multilingual QE models based on transformers trained on high-resource languages can be used for zero-shot, sentence-level QE in low-resource languages. In a similar architecture, but with multi-task learning, \citet{sun-etal-2020-exploratory} report that multilingual QE models outperform bilingual models, particularly in less balanced quality label distributions and low-resource settings. However, these two papers are focused on sentence-level QE and to the best of our knowledge, no prior work has been done on multilingual, word-level QE models.

\section{Architecture}
Our architecture relies on the XLM-R transformer model \cite{conneau-etal-2020-unsupervised} to derive the representations of the input sentences. XLM-R has been trained on a large-scale multilingual dataset in 104 languages, totalling 2.5TB, extracted from the CommonCrawl datasets. It is trained using only RoBERTa's \cite{liu2019roberta} masked language modelling (MLM) objective. XML-R was used by the winning systems in the recent WMT 2020 shared task on sentence-level QE \cite{ranasinghe-etal-2020-transquest-wmt2020, lee-2020-two,fomicheva-etal-2020-bergamot}. This motivated us to use a similar approach for word-level QE.

Our architecture adds a new token to the XLM-R tokeniser called \textsc{$<$GAP$>$} which is inserted between the words in the target. We then concatenate the source and the target with a \textsc{[SEP]} token and we feed them into XLM-R. A simple linear layer is added on top of word and {$<$GAP$>$} embeddings to predict whether it is "Good" or "Bad" as shown in Figure \ref{fig:architecture}. The training configurations and the system specifications are presented in the supplementary material.

\begin{table*}[t]
\begin{center}
\scalebox{1.00}{
\begin{tabular}{ c|c|c|c|c } 
 \hline
 \textbf{Language Pair} & \textbf{Source} & \textbf{MT System} & \textbf{Competition} &  \textbf{Train Size} \\ 
  \hline
  De-En & Pharmaceutical & Phrase-based SMT & WMT 2018 & 25,963  \\
   \hline
  En-Cs & IT & Phrase-based SMT & WMT 2018 & 40,254 \\
 \hline
    En-De & Wiki & fairseq-based NMT & WMT 2020 & 7,000 \\
  \hline
  En-De & IT & fairseq-based NMT & WMT 2019 & 13,442 \\
  \hline
  En-De & IT & Phrase-based SMT & WMT 2018 & 26,273 \\
  \hline
  En-Ru & IT & Online NMT & WMT 2019 & 15,089 \\
  \hline
  En-Lv & Pharmaceutical & Attention-based NMT & WMT 2018 & 12,936 \\
  \hline
  En-Lv & Pharmaceutical & Phrase-based SMT & WMT 2018 & 11,251   \\
\hline
   En-Zh & Wiki & fairseq-based NMT & WMT 2020 & 7,000 \\
  \hline
\end{tabular}
}
\end{center}
\caption{Information about the language pairs used to predict word-level quality. The \textbf{Language Pair} column lists the language pairs we used in ISO 639-1 codes. \textbf{Source} stands for the domain of the sentence and \textbf{MT System} is the Machine Translation system used to translate the sentences. 
\textbf{Competition} refers to the quality estimation competition in which the data was released and the last column indicates the number of instances the train dataset has for each language pair respectively.} 
\label{tab:hter_data}
\end{table*}

\section{Experimental Setup}

\subsection{QE Dataset}
We used several language pairs for which word-level QE annotations were available: English-Chinese (En-Zh), English-Czech (En-Cs), English-German (En-De), English-Russian (En-Ru), English-Latvian (En-Lv) and German-English (De-En). The texts are from a variety of domains and the translations were produced using both neural and statistical machine translation systems. More details about these datasets can be found in Table \ref{tab:hter_data} and in \citep{specia-etal-2018-findings,fonseca-etal-2019-findings, specia-etal-2020-findings-wmt}.

\renewcommand{\arraystretch}{1.2}
\begin{table*}[t]
\begin{center}
\small
\scalebox{0.77}{
\begin{tabular}{l l c c c c c c c c c} 
\toprule
& & \multicolumn{4}{c}{\bf IT} & \multicolumn{3}{c}{\bf Pharmaceutical} & \multicolumn{2}{c}{\bf Wiki}\\
\cmidrule(r){3-6}\cmidrule(r){7-9}\cmidrule(r){10-11}
&{\bf \makecell{Train \\ Language(s)} } & \makecell{En-Cs \\ SMT} & \makecell{ En-De \\ NMT} & \makecell{En-De \\ SMT} & \makecell{En-Ru \\ NMT} & \makecell{De-En \\ SMT} & \makecell{En-LV \\ NMT} & \makecell{En-Lv \\ SMT } & \makecell{En-De \\ NMT} & \makecell{En-Zh \\ NMT} \\
\midrule
\multirow{9}{*}{\bf I} & En-Cs SMT & 0.6081 & \textcolor{gray}{(-0.09)} & \textcolor{gray}{(-0.07)} & \textcolor{gray}{(-0.09)} & \textcolor{gray}{(-0.15)} & \textcolor{gray}{(-0.02)} & \textcolor{gray}{(-0.01)} & \textcolor{gray}{(-0.10)} &  \textcolor{gray}{(-0.11)}\\
& En-De NMT & \textcolor{gray}{(-0.17)} & 0.4421 & \textcolor{gray}{(-0.06)} & \textcolor{gray}{(-0.02)}  & \textcolor{gray}{(-0.18)} & \textcolor{gray}{(-0.01)} & \textcolor{gray}{(-0.02)} & \textcolor{gray}{(-0.01)} &  \textcolor{gray}{(-0.08)} \\
& En-De SMT & \textcolor{gray}{(-0.01)} & \textcolor{gray}{(-0.05)}& 0.6348 & \textcolor{gray}{(-0.67)} & \textcolor{gray}{(-0.14)} & \textcolor{gray}{(-0.06)} & \textcolor{gray}{(-0.04)} & \textcolor{gray}{(-0.06)} &  \textcolor{gray}{(-0.09)} \\
& En-Ru NMT & \textcolor{gray}{(-0.14)} & \textcolor{gray}{(-0.08)} & \textcolor{gray}{(-0.16)} & \textbf{0.5592} & \textcolor{gray}{(-0.12)} & \textcolor{gray}{(-0.01)} & \textcolor{gray}{(-0.03)} & \textcolor{gray}{(-0.09)} & \textcolor{gray}{(-0.08)} \\
& De-En SMT & \textcolor{gray}{(-0.43)} & \textcolor{gray}{(-0.23)} & \textcolor{gray}{(-0.33)} & \textcolor{gray}{(-0.31)} & \textbf{0.6485} & \textcolor{gray}{(-0.29)} & \textcolor{gray}{(-0.32)} & \textcolor{gray}{(-0.25)} & \textcolor{gray}{(-0.28)} \\
& En-LV NMT & \textcolor{gray}{(-0.12)} & \textcolor{gray}{(-0.09)} & \textcolor{gray}{(-0.14)} & \textcolor{gray}{(-0.03)} & \textcolor{gray}{(-0.12)} & 0.5868 & \textcolor{gray}{(-0.01)} & \textcolor{gray}{(0.09)} & \textcolor{gray}{(-0.08)} \\
& En-Lv SMT & \textcolor{gray}{(-0.04)} & \textcolor{gray}{(-0.16)} & \textcolor{gray}{(-0.10)} & \textcolor{gray}{(-0.09)} & \textcolor{gray}{(-0.16)} & \textcolor{gray}{(-0.01)}  & 0.5939 & \textcolor{gray}{(-0.15)} & \textcolor{gray}{(-0.14)} \\
& En-De NMT & \textcolor{gray}{(-0.11)} & \textcolor{gray}{(-0.01)} & \textcolor{gray}{(-0.08)} & \textcolor{gray}{(-0.02)} & \textcolor{gray}{(-0.14)} & \textcolor{gray}{(-0.02)}  & \textcolor{gray}{(-0.04)} & 0.5013 &  \textcolor{gray}{(-0.06)} \\
& En-Zh NMT & \textcolor{gray}{(-0.19)} & \textcolor{gray}{(-0.08)} & \textcolor{gray}{(-0.17)} & \textcolor{gray}{(-0.03)} & \textcolor{gray}{(-0.16)} & \textcolor{gray}{(-0.03)}  & \textcolor{gray}{(-0.06)} & \textcolor{gray}{(-0.07)} & 0.5402 \\
\midrule
\multirow{2}{*}{\bf II} & All & \textbf{0.6112} & \textbf{0.4523} & \textbf{0.6583} & 0.5558 & 0.6221 & \textbf{0.5991} & \textbf{0.5980} & 0.5101 & 0.5229\\
& All-1 & \textcolor{gray}{(-0.01)} & \textcolor{gray}{(-0.01)} & \textcolor{gray}{(-0.05)} & \textcolor{gray}{(-0.02)} & \textcolor{gray}{(-0.12)} & \textcolor{gray}{(-0.01)} & \textcolor{gray}{(-0.01)} & \textcolor{gray}{(-0.01)} & \textcolor{gray}{(-0.05)} \\
\midrule
\multirow{1}{*}{\bf III} & Domain & 0.6095 &  0.4467 & 0.6421 & 0.5560 & 0.6331 & 0.5892 & 0.5951  & 0.5021 &  0.5210 \\
\midrule
\multirow{1}{*}{\bf IV} & SMT/NMT & 0.6092 & 0.4461  & 0.6410 & 0.5421 & 0.6320 & 0.5885 & 0.5934 & 0.5010 & 0.5205 \\
\midrule
\multirow{3}{*}{\bf V} & Baseline-Marmot & 0.4449 & 0.1812 & 0.3630 & NR & 0.4373 & 0.4208 & 0.3445 & NR & NR \\
& Baseline-OpenKiwi & NR & NR & NR & 0.2412 & NR & NR & NR & 0.4111 & 0.5583 \\
& Best system & 0.4449 & 0.4361 & 0.6246 & 0.4780 & 0.6012 & 0.4293 & 0.3618 & \textbf{0.6186} & \textbf{0.6415} \\
\bottomrule
\end{tabular}
}
\end{center}
\caption{Target F1-Multi between the algorithm predictions and human annotations. Best results for each language by any method are marked in bold. Sections I, II and III indicate the different evaluation settings. Section IV shows the results of the state-of-the-art methods and the best system submitted for the language pair in that competition. \textbf{NR} implies that a particular result was \textit{not reported} by the organisers. Zero-shot results are coloured in grey and the value shows the difference between the best result in that section for that language pair and itself.} 
\label{tab:mt_prediction}
\end{table*}

\subsection{Evaluation Criteria}

For evaluation, we used the approach proposed in the WMT shared tasks in which the classification performance is calculated using the multiplication of F1-scores for the `OK' and `BAD' classes against the true labels
independently: words in the target (`OK' for correct words, `BAD' for incorrect words), gaps in the target (`OK' for genuine gaps, `BAD' for gaps indicating missing words) and source words (`BAD' for words that lead to errors in the target, `OK' for other words) \cite{specia-etal-2018-findings}. In recent WMT shared tasks, the most popular category was predicting quality for words in the target. 
Therefore, in Section \ref{sec:result} we only report the F1-score for words in the target. Other results are presented in the supplementary material. Prior to WMT 2019, organisers provided separate scores for gaps and words in the target, while after WMT 2019 they produce a single result for target gaps and words. We follow this latter approach.

\section{Results}
\label{sec:result}
The values displayed diagonally across section I of Table \ref{tab:mt_prediction} show the results for supervised, bilingual, word-level QE models where the model was trained on the training set of a particular language pair and tested on the test set of the same language pair. As can be seen in section V, the architecture outperforms the baselines in all the language pairs and also outperforms the majority of the best systems from previous competitions. In addition to the target word F1-score, our architecture outperforms the baselines and best systems in target gaps F1-score and source words F1-score too as shown in Tables \ref{tab:gap_prediction} and \ref{tab:source_prediction}. In the following sections we explore its behaviour in different multilingual settings.

\subsection{Multilingual QE}
We combined instances from all the language pairs and built a single word-level QE model. Our results, displayed in section II (``All'') of Table \ref{tab:mt_prediction}, show that multilingual models perform on par with bilingual models or even better for some language pairs. We also investigate whether combining language pairs that share either the same domain or MT type can be more beneficial, since it is possible that the learning process is better when language pairs share certain characteristics. However as shown in sections III and IV of Table \ref{tab:mt_prediction}, for the majority of the language pairs, specialised multilingual models built on certain domains or MT types do not perform better than multilingual models which contain all the data.

\subsection{Zero-shot QE}
To test whether a QE model trained on a particular language pair can be generalised to other language pairs, different domains and MT types, we performed zero-shot quality estimation. We used the QE model trained on a particular language pair and evaluated it on the test sets of the other language pairs. Non-diagonal values of section I in Table \ref{tab:mt_prediction} show how each QE model performed on other language pairs. 
For better visualisation, the non-diagonal values of section I of Table \ref{tab:mt_prediction} show by how much the score changes when the zero-shot QE model is used instead of the bilingual QE model. As can be seen, the scores decrease, but this decrease is negligible and is to be expected.
For most pairs, the QE model that did not see any training instances of that particular language pair outperforms the baselines that were trained extensively on that particular language pair. Further analysing the results, we can see that zero-shot QE performs better when the language pair shares some properties such as domain, MT type or language direction. For example, En-De SMT $\Rightarrow$ En-Cs SMT is better than En-De NMT $\Rightarrow$ En-Cs SMT and En-De SMT $\Rightarrow$ En-De NMT is better than En-Cs SMT $\Rightarrow$ En-De NMT.

\begin{figure*}
\centering
  \begin{subfigure}[b]{\figlength}
    \centering\includegraphics[width=\figlength]{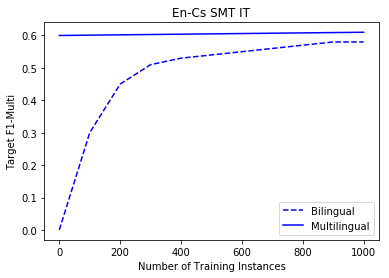}
    \caption{En-Cs SMT IT}
    \label{fig:en_cs_smt_results}
  \end{subfigure}
  \begin{subfigure}[b]{\figlength}
    \centering\includegraphics[width=\figlength]{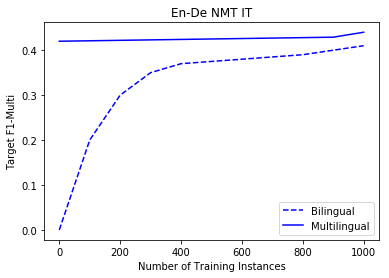}
    \caption{En-De NMT IT}
    \label{fig:en_de_nmt_it_results}
  \end{subfigure}
  \begin{subfigure}[b]{\figlength}
    \centering\includegraphics[width=\figlength]{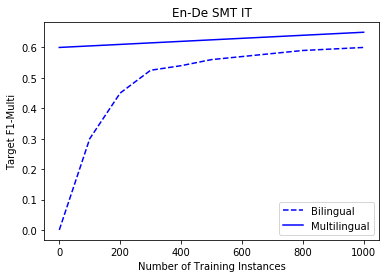}
    \caption{En-De SMT IT}
    \label{fig:en_de_smt_it_results}
  \end{subfigure}
  \begin{subfigure}[b]{\figlength}
    \centering\includegraphics[width=\figlength]{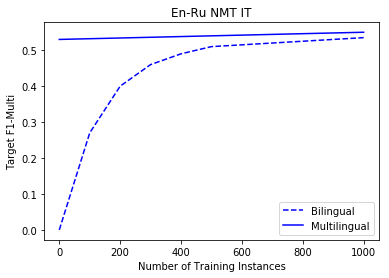}
    \caption{En-Ru NMT IT}
    \label{fig:en_ru_nmt_it_results}
  \end{subfigure}
 \begin{subfigure}[b]{\figlength}
    \centering\includegraphics[width=\figlength]{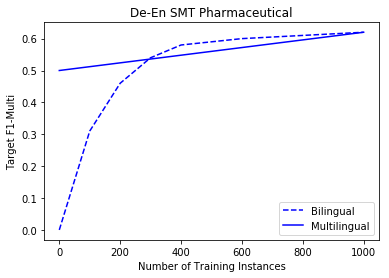}
    \caption{De-En SMT Pharmaceutical}
    \label{fig:de_en_smt_pharm_results}
  \end{subfigure}
  \begin{subfigure}[b]{\figlength}
    \centering\includegraphics[width=\figlength]{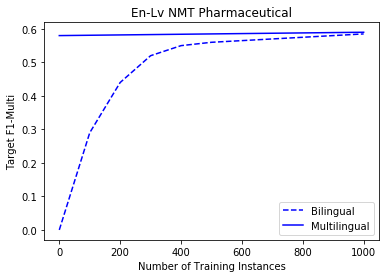}
    \caption{En-Lv NMT Pharmaceutical}
    \label{fig:en_lv_nmt_Pharm_results}
  \end{subfigure}
  \begin{subfigure}[b]{\figlength}
    \centering\includegraphics[width=\figlength]{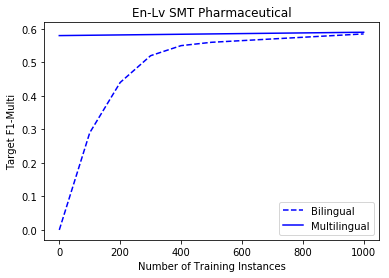}
    \caption{En-Lv SMT Pharmaceutical}
    \label{fig:en_lv_smt_pharm_results}
  \end{subfigure}
 \begin{subfigure}[b]{\figlength}
    \centering\includegraphics[width=\figlength]{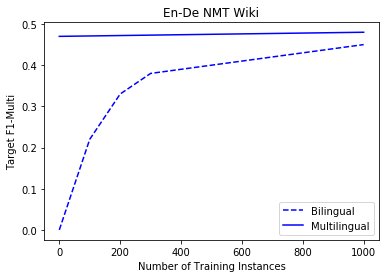}
    \caption{En-De NMT Wiki}
    \label{fig:en_de_nmt_wiki_results}
  \end{subfigure}
  \begin{subfigure}[b]{\figlength}
    \centering\includegraphics[width=\figlength]{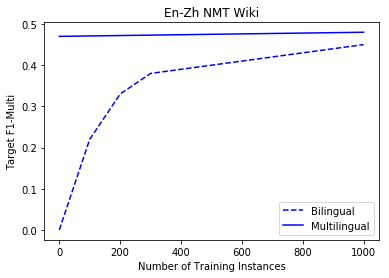}
    \caption{En-Zh NMT Wiki}
    \label{fig:en_zh_nmt_wiki_results}
  \end{subfigure}
\caption{Target F1-Multi scores with Few-shot learning}
\label{fig:fewshot_results}
\end{figure*}

We also experimented with zero-shot QE with multilingual QE models. We trained the QE model in all the pairs except one and performed prediction on the test set of the language pair left out. In section II (``All-1''), we show its difference to the multilingual QE model. This also provides competitive results for the majority of the languages, proving it is possible to train a single multilingual QE model and extend it to a multitude of languages and domains. This approach provides better results than performing transfer learning from a bilingual model.


One limitation of the zero-shot QE is its inability to perform when the language direction changes. In the scenario where we performed zero-shot learning from De-En to other language pairs, results degraded considerably from the bilingual result. 
Similarly, the performance is rather poor when we test on De-En for the multilingual zero-shot experiment as the direction of all the other pairs used for training is different. 
This is in line with results reported by \citet{ranasinghe-etal-2020-transquest} for sentence level.

\subsection{Few-shot QE}
We also evaluated how the QE models behave with a limited number of training instances. 
For each language pair, we initiated the weights of the bilingual model with those of the relevant All-1 QE and trained it on 100, 200, 300 and up to 1000 training instances. We compared the results with those obtained having trained the QE model from scratch for that language pair. The results in Figure \ref{fig:fewshot_results} show that All-1 or the multilingual model performs well above the QE model trained from scratch (Bilingual) when there is a limited number of training instances available. Even for the De-En language pair, for which we had comparatively poor zero-shot results, the multilingual model provided better results with a few training instances. It seems that having the model weights already fine-tuned in the multilingual model provides an additional boost to the training process which is advantageous in a few-shot scenario.

\section{Conclusions}
In this paper, we explored multilingual, word-level QE with transformers. We introduced a new architecture based on transformers to perform word-level QE. The implementation of the architecture, which is based on Hugging Face \cite{wolf-etal-2020-transformers}, has been integrated into the TransQuest framework \cite{ranasinghe-etal-2020-transquest} which won the WMT 2020 QE task \cite{specia-etal-2020-findings-wmt} on sentence-level direct assessment \cite{ranasinghe-etal-2020-transquest-wmt2020}\footnote{TransQuest is available on GitHub \url{https://github.com/tharindudr/TransQuest}}. In our experiments, we observed that multilingual QE models deliver excellent results on the language pairs they were trained on. In addition, the multilingual QE models perform well in the majority of the zero-shot scenarios where the multilingual QE model is tested on an unseen language pair. Furthermore, multilingual models perform very well with few-shot learning on an unseen language pair when compared to training from scratch for that language pair, proving that multilingual QE models are effective even with a limited number of training instances. While we centered our analysis around the F1-score of the target words, these findings are consistent with the F1-score of the target gaps and the F1 score of the source words too. This suggests that we can train a single multilingual QE model on as many languages as possible and apply it on other language pairs as well.
These findings can be beneficial to perform QE in low-resource languages for which the training data is scarce and when maintaining several QE models for different language pairs is arduous.

\bibliographystyle{acl_natbib}
\bibliography{acl2021}

\clearpage

\section*{Supplementary Material}
\begin{enumerate}[i]
  \item \textbf{Training Configurations}
We used an Nvidia Tesla K80 GPU to train the models. We divided the dataset into a training set and a validation set using 0.8:0.2 split. We evaluated the model while training and performed \textit{early stopping} if the validation loss did not improve over ten evaluation steps. We used the parameter values mentioned in Table \ref{tab:parameter} and did not change it across the language pairs. For all the experiments we used the XLM-R large model.

\begin{table}[!ht]
\centering
\setlength{\tabcolsep}{4.5pt}
\scalebox{0.85}{
\begin{tabular}{ll}
\hline
\bf Parameter & \bf Value  \\ \hline
learning rate & 2e-5 \\
maximum sequence length & 128 \\
number of epochs & 3 \\
adam epsilon & 1e-8       \\
warmup ratio & 0.1    \\
warmup steps  & 0       \\
max grad norm & 1.0        \\
max seq. length & 140        \\
\makecell{gradient accumulation steps} & 1 \\
 \hline
\end{tabular}
}
\caption{Parameter Specifications.}
\label{tab:parameter}
\end{table}

\item \textbf{Hardware Specifications}

In Table \ref{tab:gpu} we mention the specifications of the GPU we used for the experiments of the paper.

\begin{table}[!ht]
\centering
\setlength{\tabcolsep}{4.5pt}
\scalebox{0.85}{
\begin{tabular}{ll}
\hline
\bf Parameter & \bf Value  \\ \hline
GPU & Nvidia K80       \\
GPU Memory & 12GB    \\
GPU Memory Clock  & 0.82GHz       \\
Performance & 4.1 TFLOPS        \\
No. CPU Cores & 2        \\
RAM & 12GB        \\
 \hline
\end{tabular}
}
\caption{GPU Specifications.}
\label{tab:gpu}
\end{table}

\item \textbf{Other results}

In Table \ref{tab:gap_prediction} and in Table \ref{tab:source_prediction} we show the F1 scores for gaps in target and for words in source. They follow the same format as Table \ref{tab:mt_prediction}. The Marmot baseline used in WMT 2018 does not support quality prediction for gaps in the target and words in the source. In addition, after WMT 2019, organisers did not release scores for gaps in target. For this reason, we do not report them in Table \ref{tab:gap_prediction}.

\renewcommand{\arraystretch}{1.2}
\begin{table*}[t]
\begin{center}
\small
\scalebox{0.98}{
\begin{tabular}{l l c c c c c c c} 
\toprule
& & \multicolumn{3}{c}{\bf IT} & \multicolumn{3}{c}{\bf Pharmaceutical} \\
\cmidrule(r){3-5}\cmidrule(r){6-8}
&{\bf \makecell{Train \\ Language(s)} } & \makecell{En-Cs \\ SMT} & \makecell{ En-De \\ NMT} & \makecell{En-De \\ SMT} & \makecell{De-En \\ SMT} & \makecell{En-LV \\ NMT} & \makecell{En-Lv \\ SMT }  \\
\midrule
\multirow{9}{*}{\bf I} & En-Cs SMT & 0.2018 & \textcolor{gray}{(-0.10)} & \textcolor{gray}{(-0.08)}  & \textcolor{gray}{(-0.15)} & \textcolor{gray}{(-0.02)} & \textcolor{gray}{(-0.01)} \\
& En-De NMT & \textcolor{gray}{(-0.17)} & 0.1672 & \textcolor{gray}{(-0.07)}  & \textcolor{gray}{(-0.18)} & \textcolor{gray}{(-0.01)} & \textcolor{gray}{(-0.02)} \\
& En-De SMT & \textcolor{gray}{(-0.08)} & \textcolor{gray}{(-0.05)}& 0.4927  & \textcolor{gray}{(-0.14)} & \textcolor{gray}{(-0.06)} & \textcolor{gray}{(-0.04)} \\
& En-Ru NMT & \textcolor{gray}{(-0.14)} & \textcolor{gray}{(-0.00)} & \textcolor{gray}{(-0.15)}  & \textcolor{gray}{(-0.12)} & \textcolor{gray}{(-0.01)} & \textcolor{gray}{(-0.03)} \\
& De-En SMT & \textcolor{gray}{(-0.18)} & \textcolor{gray}{(-0.14)} & \textcolor{gray}{(-0.33)}  & \textbf{0.4203} & \textcolor{gray}{(-0.29)} & \textcolor{gray}{(-0.32)}  \\
& En-LV NMT & \textcolor{gray}{(-0.16)} & \textcolor{gray}{(-0.09)} & \textcolor{gray}{(-0.15)}  & \textcolor{gray}{(-0.12)} & 0.1664 & \textcolor{gray}{(-0.01)} \\
& En-Lv SMT & \textcolor{gray}{(-0.11)} & \textcolor{gray}{(-0.12)} & \textcolor{gray}{(-0.11)}  & \textcolor{gray}{(-0.16)} & \textcolor{gray}{(-0.01)}  & 0.2356 \\
& En-De NMT & \textcolor{gray}{(-0.17)} & \textcolor{gray}{(-0.01)} & \textcolor{gray}{(-0.09)}  & \textcolor{gray}{(-0.14)} & \textcolor{gray}{(-0.02)}  & \textcolor{gray}{(-0.04)} \\
& En-Zh NMT & \textcolor{gray}{(-0.15)} & \textcolor{gray}{(-0.08)} & \textcolor{gray}{(-0.16)} & \textcolor{gray}{(-0.16)} & \textcolor{gray}{(-0.03)}  & \textcolor{gray}{(-0.06)} \\
\midrule
\multirow{2}{*}{\bf II} & All & \textbf{0.2118} & \textbf{0.1773} & \textbf{0.5028} &  0.4189 & \textbf{0.1772} & \textbf{0.2388} \\
& All-1 & \textcolor{gray}{(-0.03)} & \textcolor{gray}{(-0.04)} & \textcolor{gray}{(-0.08)} &  \textcolor{gray}{(-0.14)} & \textcolor{gray}{(-0.01)} & \textcolor{gray}{(-0.01)} \\
\midrule
\multirow{1}{*}{\bf III} & Domain & 0.2112 &  0.1695 & 0.4951 &  0.4132 & 0.1685 & 0.2370  \\
\midrule
\multirow{1}{*}{\bf IV} & SMT/NMT & 0.2110 & 0.1886  & 0.4921 &  0.4026 & 0.1671 & 0.2289 \\
\midrule
\multirow{2}{*}{\bf V} & Marmot & 0.0000 & 0.0000 & 0.0000  & 0.0000 & 0.0000 & 0.0000 \\
& Best system & 0.1671 & 0.1343 & 0.3161 & 0.3176 & 0.1598 & 0.1386 \\
\bottomrule
\end{tabular}
}
\end{center}
\caption{GAP F1-Multi between the algorithm predictions and human annotations. Best results for each language by any method are marked in bold. Sections I, II and III indicate the different evaluation settings. Section IV shows the results of the state-of-the-art methods and the best system submitted for the language pair in that competition. \textbf{NR} implies that a particular result was \textit{not reported} by the organisers. Zero-shot results are coloured in grey and the value shows the difference between the best result in that column for that language and itself.} 
\label{tab:gap_prediction}
\end{table*}

\renewcommand{\arraystretch}{1.2}
\begin{table*}[t]
\begin{center}
\small
\scalebox{0.95}{
\begin{tabular}{l l c c c c c c c c c} 
\toprule
& & \multicolumn{4}{c}{\bf IT} & \multicolumn{3}{c}{\bf Pharmaceutical} & \multicolumn{2}{c}{\bf Wiki}\\
\cmidrule(r){3-6}\cmidrule(r){7-9}\cmidrule(r){10-11}
&{\bf \makecell{Train \\ Language(s)} } & \makecell{En-Cs \\ SMT} & \makecell{ En-De \\ NMT} & \makecell{En-De \\ SMT} & \makecell{En-Ru \\ NMT} & \makecell{De-En \\ SMT} & \makecell{En-LV \\ NMT} & \makecell{En-Lv \\ SMT } & \makecell{En-De \\ NMT} & \makecell{En-Zh \\ NMT} \\
\midrule
\multirow{9}{*}{\bf I} & En-Cs SMT & 0.5327 & \textcolor{gray}{(-0.08)} & \textcolor{gray}{(-0.07)} & \textcolor{gray}{(-0.09)} & \textcolor{gray}{(-0.17)} & \textcolor{gray}{(-0.02)} & \textcolor{gray}{(-0.01)} & \textcolor{gray}{(-0.12)} &  \textcolor{gray}{(-0.13)}\\
& En-De NMT & \textcolor{gray}{(-0.17)} & 0.2957 & \textcolor{gray}{(-0.07)} & \textcolor{gray}{(-0.02)}  & \textcolor{gray}{(-0.19)} & \textcolor{gray}{(-0.01)} & \textcolor{gray}{(-0.02)} & \textcolor{gray}{(-0.02)} &  \textcolor{gray}{(-0.08)} \\
& En-De SMT & \textcolor{gray}{(-0.01)} & \textcolor{gray}{(-0.05)}& 0.5269 & \textcolor{gray}{(-0.67)} & \textcolor{gray}{(-0.14)} & \textcolor{gray}{(-0.06)} & \textcolor{gray}{(-0.05)} & \textcolor{gray}{(-0.08)} &  \textcolor{gray}{(-0.09)} \\
& En-Ru NMT & \textcolor{gray}{(-0.14)} & \textcolor{gray}{(-0.08)} & \textcolor{gray}{(-0.18)} & \textbf{0.5543} & \textcolor{gray}{(-0.14)} & \textcolor{gray}{(-0.01)} & \textcolor{gray}{(-0.03)} & \textcolor{gray}{(-0.09)} & \textcolor{gray}{(-0.08)} \\
& De-En SMT & \textcolor{gray}{(-0.42)} & \textcolor{gray}{(-0.21)} & \textcolor{gray}{(-0.33)} & \textcolor{gray}{(-0.31)} & \textbf{0.4824} & \textcolor{gray}{(-0.29)} & \textcolor{gray}{(-0.32)} & \textcolor{gray}{(-0.23)} & \textcolor{gray}{(-0.28)} \\
& En-LV NMT & \textcolor{gray}{(-0.12)} & \textcolor{gray}{(-0.09)} & \textcolor{gray}{(-0.14)} & \textcolor{gray}{(-0.03)} & \textcolor{gray}{(-0.12)} & 0.4880 & \textcolor{gray}{(-0.01)} & \textcolor{gray}{(0.09)} & \textcolor{gray}{(-0.08)} \\
& En-Lv SMT & \textcolor{gray}{(-0.04)} & \textcolor{gray}{(-0.16)} & \textcolor{gray}{(-0.11)} & \textcolor{gray}{(-0.09)} & \textcolor{gray}{(-0.17)} & \textcolor{gray}{(-0.02)}  & 0.4945  & \textcolor{gray}{(-0.15)} & \textcolor{gray}{(-0.14)} \\
& En-De NMT & \textcolor{gray}{(-0.11)} & \textcolor{gray}{(-0.01)} & \textcolor{gray}{(-0.08)} & \textcolor{gray}{(-0.02)} & \textcolor{gray}{(-0.15)} & \textcolor{gray}{(-0.03)}  & \textcolor{gray}{(-0.04)} & 0.4456 &  \textcolor{gray}{(-0.06)} \\
& En-Zh NMT & \textcolor{gray}{(-0.19)} & \textcolor{gray}{(-0.08)} & \textcolor{gray}{(-0.17)} & \textcolor{gray}{(-0.03)} & \textcolor{gray}{(-0.18)} & \textcolor{gray}{(-0.05)}  & \textcolor{gray}{(-0.06)} & \textcolor{gray}{(-0.07)} & 0.4040 \\
\midrule
\multirow{2}{*}{\bf II} & All & \textbf{0.5442} & \textbf{0.3021} & \textbf{0.5445} & 0.5535 & 0.4791 & \textbf{0.4983} & \textbf{0.5005} & 0.4483 & 0.4053\\
& All-1 & \textcolor{gray}{(-0.02)} & \textcolor{gray}{(-0.02)} & \textcolor{gray}{(-0.06)} & \textcolor{gray}{(-0.03)} & \textcolor{gray}{(-0.16)} & \textcolor{gray}{(-0.01)} & \textcolor{gray}{(-0.01)} & \textcolor{gray}{(-0.01)} & \textcolor{gray}{(-0.04)} \\
\midrule
\multirow{1}{*}{\bf III} & Domain & 0.5421 &  0.2925 & 0.5421 & 0.5259 & 0.4672 & 0.4907 & 0.4991  & 0.4364 &  0.4021 \\
\midrule
\multirow{1}{*}{\bf IV} & SMT/NMT & 0.5412 & 0.2901  & 0.5412 & 0.5230 & 0.4670 & 0.4889 & 0.4932 & 0.4302 & 0.4012 \\
\midrule
\multirow{3}{*}{\bf V} & Marmot & 0.0000 & 0.0000 & 0.0000 & NR & 0.0000 & 0.0000 & 0.0000 & NR & NR \\
& OpenKiwi & NR & NR & NR & 0.2647 & NR & NR & NR & 0.3717 & 0.3729 \\
& Best system & 0.3937 & 0.2642 & 0.3368  & 0.4541 & 0.3200 & 0.3614 & 0.4945 & \textbf{0.5672} & \textbf{0.4462} \\
\bottomrule
\end{tabular}
}
\end{center}
\caption{SOURCE F1-Multi between the algorithm predictions and human annotations. Best results for each language by any method are marked in bold. Rows I, II and III indicate the different evaluation settings. Row IV shows the results of the state-of-the-art methods and the best system submitted for the language pair in that competition. \textbf{NR} implies that a particular result was \textit{not reported} by the organisers. Zero-shot results are coloured in grey and the value shows the difference between the best result in that column for that language and itself.} 
\label{tab:source_prediction}
\end{table*}

\end{enumerate}

\end{document}